\documentclass[conference]{IEEEtran}

\usepackage[pdftex]{graphicx}
\usepackage{url}
\usepackage{caption}
\usepackage{graphicx}
\usepackage{amssymb}
\usepackage{amsthm}

\newcommand\dep{\mathcal{D}}
\newcommand\ar{\mathcal{A}}
\newcommand\B{\mathcal{B}}
\newcommand\e{\mathcal{E}}
\newcommand\x{\textit{X}}
\newcommand\y{\textit{Y}}

\usepackage{times}

\def\Cons{\textsf{\textbf{C}}}
\def\actors{\Cons_{act}}
\def\ops{\Cons_{exp}}
\def\Var{\textsf{\textbf{{V}}}}
\def\PredA{\textsf{\textbf{{P}}}}
\def\GndA{\textsf{\textbf{{G}}}}

\newcommand{\upperRomannumeral}[1]{\uppercase\expandafter{\romannumeral#1}}
\newtheorem{example}{Example}
\newtheorem{definition}{Definition}

\newcommand{\Tree}[1]{\mbox{${\mathcal T}({\small #1})$}}
\newcommand{\MTree}[1]{\mbox{${\mathcal T^*}({\small #1})$}}

\IEEEoverridecommandlockouts

\begin{document}
	
	\renewcommand{\thetable}{\arabic{table}}
	\renewcommand{\figurename}{Fig.}

\title{Argumentation Models for Cyber Attribution}

\author{\IEEEauthorblockN{Eric Nunes, Paulo Shakarian}
	\IEEEauthorblockA{
		Arizona State University\\
		Tempe, AZ 85281, USA\\
		Email: \{enunes1, shak\} @asu.edu}
	\and
	\IEEEauthorblockN{Gerardo I. Simari}
	\IEEEauthorblockA{Inst.\ for CS and Eng.\ (CONICET--UNS)\\
		DCIC, UNS, Bahia Blanca, Argentina\\
		Email: gis@cs.uns.edu.ar}
	\and 
	\IEEEauthorblockN{Andrew Ruef}
	\IEEEauthorblockA{Trail of Bits, Inc.\\
		New York, NY 10003, USA\\
		Email: andrew@trailofbits.com}
	}

\maketitle

\begin{abstract}
A major challenge in cyber-threat analysis is combining information from different sources to find the person or the group responsible for the cyber-attack. It is one of the most important technical and policy challenges in cyber-security. The lack of ground truth for an individual responsible for an attack has limited previous studies. In this paper, we take a first step towards overcoming this limitation by building a dataset from the capture-the-flag event held at DEFCON, and propose an argumentation model based on a formal reasoning framework called DeLP (Defeasible Logic Programming) designed to aid an analyst in attributing a cyber-attack. We build models from latent variables to reduce the search space of culprits (attackers), and show that this reduction significantly improves the performance of classification-based approaches from 37\% to 62\% in identifying the attacker.
\end{abstract}

\IEEEpeerreviewmaketitle

\section{Introduction}
A major challenge in cyber-threat analysis is to find the person or the group responsible for a cyber-attack. This is known as cyber-attribution~\cite{warfare} and it is one of the central technical and policy challenges in cyber-security. Oftentimes, the evidence collected from multiple sources provides a contradictory viewpoint. This gets worse in cases of deception where either an attacker plants false evidence or the evidence points to multiple actors, leading to uncertainty. In the text on cyber-warfare~\cite{warfare} the authors discuss the difficulties that an intelligence analyst faces in attributing an attack to a perpetrator given that deception might have occurred, and how the analyst needs to explore deception hypotheses under the given attack scenario.

However, one of the major drawbacks of the study and evaluation of cyber-attribution models is the lack of datasets with the ground truth available regarding the individual party responsible for the attack---this has limited previous studies. To overcome this, we built and leveraged a dataset from the capture-the-flag event held at DEFCON. In previous work, this dataset was used to study cyber-attribution, framing it as a multi-label classification problem to predict the attacker~\cite{fosint}. Machine learning approaches struggle in situations of {\em deception}, where similar attributes point towards multiple attackers---we propose to address this issue using a formal logical framework. 

Specific contributions of this paper include:
\begin{itemize}
	\setlength{\itemsep}{0em}
	\item description of how a model for cyber-attribution can be designed and implemented in the DeLP structured argumentation framework;
	\item experiments demonstrating that using argumentation-based tools can significantly reduce the number of potential culprits that need to be considered in the analysis of a cyber-attack; and
	\item experiments showing that the reduced set of culprits, used in conjunction with classification, leads to improved cyber-attribution decisions.
	
\end{itemize}

\noindent\textbf{Related work:} 
Adversarial machine learning is an emerging field of study. It uses effective machine learning techniques to identify or defend against an adversary's opponents. Understanding the limits of adversary's knowledge and capabilities is crucial for coming up with countermeasures, as discussed in~\cite{huang2011adversarial}. Here the authors propose models to study these limitations to come up with evasion techniques. On the contrary, Lowd and Meek~\cite{lowd2005adversarial} explore the problem from an adversarial point of view. They propose strategies that an adversary can use to reverse engineer a classifier so that his attacks are undetected by the classifier. They use a real world application in spam filtering to demonstrate their method, which they call adversarial classifier evasion. In a spam filtering setting an example of such a technique is replacing feature words that raise a red flag with their synonyms to evade detection. This feature cross substitution technique is discussed in~\cite{NIPS2014_5510}. Here the authors offer a simple heuristic method based on  mixed-integer linear programming with constraint generation to make the classifier robust to cross substitution techniques. There is research that looks at modeling the interaction between the learner (adversary) and the classifier in terms of a competition using Stackelberg games~\cite{bruckner2011stackelberg,bruckner2012static}. Most adversarial machine learning applications deal with modeling classifiers to be robust against evasive techniques in real world applications like malware detection and spam filtering. Cyber-attribution falls in the domain of adversarial learning, but looks at analyzing the evidence in the aftermath of an attack to discover the attacker. 

Currently, cyber-attribution is limited to identifying machines~\cite{survey} as opposed to the hacker or their affiliation to a group or a state. An example of such a technical attribution approach is WOMBAT~\cite{wombat}, where a clustering technique is used to group attacks to common IP sources. A method that combines information from different sources was proposed by Walls~\cite{walls}, who considered forensic information from diverse sources but did not account for inconsistency or uncertainty due to deception. A less rigorous mathematical model, known as the Q model~\cite{Q}, was proposed recently; the model answers queries from an analyst,  
and by combining these answers the analyst attributes an attack to a party. Unfortunately, there are no experimental evaluations of its effectiveness. Argumentation has been used for cyber reasoning~\cite{applebaumcyber} by leveraging arguments to deal with incomplete and contradictory data, allowing to derive big-picture conclusions to keep systems secure and online in case of an attack. This is a different application than the one we are addressing. 

In~\cite{sklar2015evaluation}, a tool was presented to support human decisions, focusing on how user trust in the evidence influences the process; a user study demonstrating the hypotheses was presented in~\cite{salvit2014employing}. Concurrently, a formal logical framework for reasoning about cyber-attribution has been devised~\cite{shakarian,shakarian14}; it explores multiple competing hypotheses based on the evidence for and against a particular attacker to help analysts decide on an attribution, providing a map of the reasoning that led to the decision. 

The rest of the paper is organized as follows. We present a description of our DEFCON capture-the-flag dataset and an analysis on the occurrence of deception within this data in Section~\ref{dataset}. This is followed by the argumentative model based on~\cite{garcia2004} in Section~\ref{arg}. We then summarize results from~\cite{fosint} and discuss how we built our baseline argumentation model along with two other extended baseline models for cyber-attribution with DeLP in Section~\ref{eb1} and Section~\ref{eb2}, with a discussion of the experimental results obtained with each of these models. Conclusions are discussed in Section~\ref{con}.

\section{DEFCON CTF Dataset}
\label{dataset}
The DEFCON security conference sponsors and hosts a capture the flag (CTF) 
competition every year, held on site with the conference in Las Vegas, Nevada. DEFCON
CTF is one of the oldest and best-known competitions. The ctftime.org site
provides a ranking for CTF teams and CTF competitions, and in this system
DEFCON CTF has the highest average weight of all other CTF competitions. 

CTF competitions can be categorized by what role the competitors play in the 
competition: either red team, blue team, or a combination. In a blue team 
focused CTF the competitors harden their systems against a red team played
by the organizers of the CTF. In a combined red/blue team CTF every team plays
both blue and red team simultaneously. The NCCDC and CDX competitions are 
examples of a blue team CTF, while DEFCON CTF is a combined red/blue team. 
Each team is simultaneously responsible for hardening and defending their 
systems as well as identifying vulnerabilities and exploiting them in 
other teams' systems. 

The game environment is created primarily by the 
DEFCON CTF organizers. The game focuses around programs (known
in the game as \textit{services}) written by the organizers. These services are 
engineered to contain specific vulnerabilities. The binary image of the 
service is made available to each team at the start of the game, but no 
other information about the service is released. Part of the challenge
of the game is identifying the purpose of each service as well as the 
vulnerabilities present in the service. Identification of vulnerabilities
serves both a defensive and offensive goal. Once a vulnerability has been
identified, a team may patch this vulnerability in the binary program. 
Additionally, the teams may create exploits for that vulnerability and use
them to attack other teams and capture digital flags from those teams' systems. 

Each team is also provided with a server running the services, which contains the digital flags to be defended. To deter defensive actions 
such as powering off the server or stopping the services, the white team (a third team, played by the organizers)
conducts periodic availability tests of the services running on each team's 
server. A team's score is the sum of the value of the flags they have 
captured, minus the sum of the flags that have been captured from that team, 
multiplied by an availability score determined by how often the white team was
able to test that team's services. This scoring model incentivizes teams to 
keep their server online, identify the vulnerabilities in services and 
patch them quickly, and exploit other teams services to capture their flags. 
It disincentivizes teams' from performing host-level blocking and shutting down
services, as this massively impacts their final score. 

This game environment can be viewed as a microcosm of the global Internet 
and the careful game of cat and mouse between hacking groups and companies. 
Teams are free to use different technical means to discover vulnerabilities. 
They may use fuzzing and reverse engineering on their own programs, or, they 
may monitor the network data sent to their services and dynamically study the 
effects that network data has on unpatched services. If a team discovers a 
vulnerability and uses it against another team, the first team may discover 
that their exploit is re-purposed and used against them within minutes.

The organizers of DEFCON CTF capture all of the network traffic sent and
received by each team, and publish this traffic at the end of the competition~\cite{pcap}.
This includes IP addresses for source and destination, as well as the full
data sent and received and the time the data was sent or received. This
data is not available to contestants; depending on the organizers' choice
from year to year, the contestants either have a real time feed but with the
IP address obscured, or a full feed delivered on a time delay of minutes
to hours.

\noindent\textbf{Analysis:}
We use the data from the CTF tournament held at DEFCON~21 in 2013.
The CTF data set is very large, about 170 GB in compressed format. We used multiple
systems with distributed and coordinated processing to analyze the entire
dataset---fortunately, analyzing individual streams is easy to parallelize. To analyze this data, we identified the TCP ports
associated with each vulnerable service. From this information, we used
the open source tool \textit{tcpflow} to process the network captures into a set of files, with each file representing data sent or received on a particular connection.

With these data files identified, we analyzed some of them by hand using the
Interactive Disassembler (IDA) to determine if the data contained shell-code,
which in fact was the case. We used an automated tool to produce a summary of each data file
as a JSON encoded element. Included in this summary was a hash of the contents
of the file and a histogram of the processor instructions contained in the file.
These JSON files were the final output of the low-level analysis, transforming
hundreds of gigabytes of network traffic into a manageable set of facts about
exploit traffic in the data. Each JSON file is a list of tuples (time-stamp, byte-histogram, instruction-histogram, attack team and target team). The individual fields of the tuple are listed in Table~\ref{exAt}. 

\begin{table}[t]
	\caption{\textmd{Fields in an instance of network attack}}
	\footnotesize
	\label{exAt}
	\centering
	\renewcommand{\arraystretch}{1.5}
	
	\begin{tabular}{|p{1.4cm}|p{3.4cm}|p{2.6cm}|}
		\hline
		{\bf Field} &  {\bf Intuition} & {\bf Value} \\ \hline

		\textsf{byte\_hist} & Histogram of byte sequences in the payload & $0{\times}43$:245, $0{\times}69$:8, $0{\times}3a$:9, .....\\ \hline
		\textsf{inst\_hist} & Histogram of instructions used in the payload & cmp:12, subs:8, movtmi:60 ...... \\ \hline
		\textsf{from\_team} & The team where the payload originates (attacking team) & Blue Lotus \\ \hline
		\textsf{to\_team} & The team being attacked by the exploit & Robot Mafia\\ \hline
		\textsf{time} & Indicates the date and time of the attack & 2013-08-03T23:45:17\\
		\hline
	\end{tabular}

\end{table}

The pre-processing can be summarized in the following steps:

\begin{itemize}
	\setlength{\itemsep}{0em}
	\item Un-tarring the archives available from the organizers. The archives produce a large number of pcap-ng formatted files that contain the traffic captures.
	\item Conversion of the pcap-ng files to tcpdump format capture using the editcap utility. This is to allow tcpflow to process the data. 
	\item  Use of xargs and GNU parallel to run tcpflow on each pcap. This is a time-consuming process, and produced a directory structure with files for data sent and received on host-port socket pairs. This step of processing allows file-based tools to process the network data.
	\item A tool to process each file containing data sent or received by network ports associated with CTF challenges. These tools produced summary statistics for each data stream, to include a byte histogram, overall size, a hash, and an ARM instruction histogram (we ran a linear sweep with the Capstone instruction decoder to produce this). This data was saved via JSON.
\end{itemize}

After this pre-processing of the network data packets, we have around 10 million network attacks consisting of around 1 million unique exploits built and used by 20 teams in the competition. In order to attribute an attack to a particular team, apart from analyzing the payloads used by the team, we also need to analyze the behavior of the attacking team towards their adversary. For this purpose, we divide the attacks according to the team being targeted. Thus, we have 20 such subsets, which we represent as T-$i$, where $i \in \{1, 2, 3, ..., 20\}$.
The processed dataset is publicly available \footnote{http://lab.engineering.asu.edu/cysis/cyber-attribution/}.

We now discuss two important observations from the dataset, which make the task of attributing an observed network attack to a team difficult.

\noindent\textbf{Deception:}
In the context of this paper we define an attack to be deceptive when multiple adversaries get mapped to a single attack pattern; {\em deception} is thus a scenario in which the same exploit is used by multiple teams to target the same team. The number of unique deceptive attacks amount to just under 35\% of the total unique attacks in our dataset---clearly, deception is a heavily-used technique in this domain.

\noindent\textbf{Duplicate attacks:}
A duplicate attack occurs when the same team uses the same payload to attack the same team at different points in time. We group duplicates as either being {\em non-deceptive} or {\em deceptive}. Non-deceptive duplicates are the copies of the attacks launched by the team that first initiated the use of a particular payload; on the other hand, deceptive duplicates are all the attacks from the teams that did not initiate the use. 

\section{Argumentation Model}
\label{arg}
Our approach relies on a model of the world where we can analyze competing hypotheses in a cyber-operation scenario. Such a model should allow for contradictory information so it can handle inconsistency in cases of deception.

Before describing the argumentation model in detail, we introduce some necessary notation. Variables and constant symbols represent items such as the exploits/payloads used for the attack, and the actors conducting the cyber-attack (in this case, the teams in the CTF competition). We denote the set of all variable symbols with $\Var$ and the set of all constants with $\Cons$. For our model we require two subsets of $\Cons$: $\actors$, denoting the actors capable of conducting the cyber-operation, and $\ops$, denoting the set of unique exploits used. We use symbols in all capital letters to denote variables. In the running example, we use a subset of our DEFCON CTF dataset.

\begin{example}
	Actors and cyber-operations from the CTF data:
	$\actors = \{\textit{bluelotus, robotmafia, apt8}\}$,
	$\ops = \{\textit{exploit}_1, \textit{exploit}_2, ..., \textit{exploit}_n\}$.
\end{example}

The language also contains a set of predicate symbols that have constants or variables as arguments, and denote events that can be either \textit{true} or \textit{false}. We denote the set of predicates with $\PredA$; examples of predicates are shown in Table~\ref{example}. For instance, $\textit{culprit}(\textit{exploit}_1,\textit{apt8})$ will either be true or false, and denotes the event where $\textit{apt8}$ used $\textit{exploit}_1$ to conduct a cyber-operation.

\vspace{-1em}
\begin{table}[ht]
	\caption{\textmd{Example predicates and explanation}}
	\footnotesize
	\label{example}
	\centering
	\renewcommand{\arraystretch}{1.5}
	
	\begin{tabular}{|p{3.3cm}|p{4cm}|}
		\hline
		{\bf Predicate} &  {\bf Explanation} \\ \hline
		
		\mbox{\textsf{attack}$(\textit{exploit}_1$, \textit{bluelotus})} & $\textit{exploit}_1$ was targeted towards the team Blue Lotus.\\ \hline
		\mbox{\textsf{replay\_attack}$(\e,\y)$} & Exploit $\e$ was replayed by team~$\y$.\\ \hline
		\mbox{\textsf{deception}$(\textit{exploit}_1, apt8)$} & Team $\textit{apt8}$ used $\textit{exploit}_1$ for deception.\\ \hline
		\mbox{\textsf{time\_diff}$(I, \y)$} & Team $\y$ was deceptive within the given time interval $I$. \\ \hline
		\mbox{\textsf{culprit}$(\textit{exploit}_1, \textit{apt8})$} & Team $\textit{apt8}$ is the likely culprit for the attack (using $\textit{exploit}_1$ on the target team).\\

		\hline
	\end{tabular}
\end{table}

A ground atom is composed by a predicate symbol and a tuple of constants, one for each argument. The set of all ground atoms is denoted as $\GndA$. A ground literal $L$ is a ground atom or a negated ground atom; hence, ground literals have no variables.
An example of a ground atom for our running example is $\textit{attack}(\textit{exploit}_1, \textit{bluelotus})$.
We denote a subset of $\GndA$ with $\GndA'$.

We choose a structured argumentation framework~\cite{rahwan2009} for our model; our approach works by creating arguments (in the form of a set of rules and facts) that compete with each other to attribute an attack to a given perpetuator. In this case, arguments are defeated based on contradicting information in other arguments. This procedure is known as a {\em dialectical process}, where the arguments that are undefeated prevail. An important result is the set of all the arguments that are \textit{warranted} (not defeated) by any other argument, which give a clear map supporting the conclusion.
Such transparency lets a security analyst not only add new arguments based on new evidence discovered in the system, but also get rid of incorrect information and fine-tune the model for better performance. Since the argumentation model can deal with inconsistent information, it draws a natural analogy to the way humans settle disputes when there is contradictory information available. Having a clear explanation of why one argument is chosen over others is a desirable characteristic for both the analyst and for organizations to make decisions and policy changes. We now briefly discuss some preliminaries on DeLP.

\noindent\textbf{Defeasible Logic Programming:}
DeLP is a formalism that combines logic programming  with defeasible argumentation; full details are discussed in~\cite{garcia2004}. The formalism is made up of several constructs, namely facts, strict rules, and defeasible rules. Facts represent statements obtained from evidence, and are always true; similarly, strict rules are logical combinations of elements (facts or other inferences) that can always be performed. On the contrary, defeasible rules can be thought of as strict rules that may be true in some situations, but could be false if contradictory evidence is present. These three constructs are used to build arguments, and DeLP programs are sets of facts, strict rules and defeasible rules. We use the usual notation for DeLP programs, denoting the knowledge base with $\Pi = (\Theta, \Omega, \Delta)$, where $\Theta$ is the set of facts, $\Omega$ is the set of strict rules, and $\Delta$ is the set of defeasible rules. Examples of the three constructs are provided with respect to the dataset in \figurename~\ref{fig:gndArgEx}. We now describe the constructs in detail.

\noindent\textbf{Facts ($\Theta$)} are ground literals that represent atomic information or its (strong) negation ($\neg$).

\smallskip
\noindent\textbf{Strict Rules ($\Omega$)} represent cause and effect information; they are of the form 
$L_0 \leftarrow L_1,...L_n$, where $L_0$ is a literal and $\{L_i\}_{i>0}$ is a set of literals.

\smallskip
\noindent\textbf{Defeasible Rules ($\Delta$)} are weaker versions of strict rules, and are of the form 
$L_0$ -$\!\prec$ $L_1,....,L_n$, where $L_0$, is the literal and $\{L_i\}_{i>0}$ is a set of literals.

When a cyber-attack occurs, the model can be used to derive arguments as to who could have conducted the attack. Derivation follows the same mechanism as logic programming~\cite{lloyd2012}.
DeLP incorporates defeasible argumentation, which decides which arguments are warranted and it blocks arguments that are in conflict and a winner cannot be determined.
\figurename~\ref{fig:gndArgEx} shows a ground argumentation framework demonstrating constructs derived from the CTF data. For instance, $\theta_1$ indicates the fact that $\textit{exploit}_1$ was used to target the team \textit{Blue Lotus}, and $\theta_5$ indicates that team \textit{pwnies} is the most frequent user of $\textit{exploit}_1$. For the strict rules, $\omega_1$ says that for a given $\textit{exploit}_1$ the attacker is \textit{pwnies} if it was the most frequent attacker and the attack $\textit{exploit}_1$ was replayed. Defeasible rules can be read similarly; $\delta_2$ indicates that $\textit{exploit}_1$ was used in a deceptive attack by APT8 if it was replayed and the first attacker was not APT8. By replacing the constants with variables in the predicates we can derive a non-ground argumentation framework.

\begin{figure}[t]
	\small
	\fbox{
		\parbox{0.96\columnwidth}{
			\begin{tabular}{lll}
				\small
				$\Theta:$& $\theta_{1}=$ & \textsf{attack}$(\textit{exploit}_1, \textit{bluelotus})$ \\
				& $\theta_{2}=$ & \textsf{first\_attack}$(\textit{exploit}_1, \textit{robotmafia})$ \\
				& $\theta_3=$ & \textsf{last\_attack}$(\textit{exploit}_1, \textit{apt8}))$ \\
				& $\theta_4=$ & \textsf{time\_diff}$(\textit{interval}, \textit{robotmafia})$ \\
				& $\theta_5=$ & \textsf{most\_frequent}$(\textit{exploit}_1, \textit{pwnies})$ \\
				\multicolumn{3}{c}{\rule{0.9\columnwidth}{0.4pt}} \\
				$\Omega:$& $\omega_{1}=$ & \textsf{culprit}$(\textit{exploit}_1, \textit{pwnies}) \leftarrow$ \\
				&            &  \hskip6mm \textsf{most\_frequent}$(\textit{exploit}_1, \textit{pwnies})$,\\
				&           & \hskip6mm \textsf{replay\_attack}$(\textit{exploit}_1)$ \\
				&$\omega_{2}=$ & \textsf{$\neg$ culprit}$(\textit{exploit}_1, \textit{robotMafia}) \leftarrow$ \\
				&            &  \hskip6mm \textsf{last\_attack}$(\textit{exploit}_1, \textit{apt8})$,\\
				&           & \hskip6mm \textsf{replay\_attack}$(\textit{exploit}_1)$ \\
				\multicolumn{3}{c}{\rule{0.9\columnwidth}{0.4pt}} \\
				$\Delta:$& $\delta_{1}=$ & \textsf{replay\_attack}$(\textit{exploit}_1)$ -$\!\prec$ \\
				&            & \hskip6mm  \textsf{attack}$(\textit{exploit}_1, \textit{bluelotus})$, \\
				&            & \hskip6mm  \textsf{last\_attack}$(\textit{exploit}_1, \textit{apt8})$\\
				&$\delta_{2}=$ & \textsf{deception}$(\textit{exploit}_1, \textit{apt8})$ -$\!\prec$  \\
				&            & \hskip6mm  \textsf{replay\_attack}$(\textit{exploit}_1)$, \\
				&            & \hskip6mm  \textsf{first\_attack}$(\textit{exploit}_1, \textit{robotmafia})$ \\
				&$\delta_{3}=$ & \textsf{culprit}$(\textit{exploit}_1, \textit{apt8})$ -$\!\prec$ \\
				&            & \hskip6mm  \textsf{deception}$(\textit{exploit}_1, \textit{apt8})$,  \\
				&            & \hskip6mm  \textsf{replay\_attack}$(\textit{exploit}_1)$ \\
				&$\delta_{4}=$ & \textsf{$\neg$culprit}$(\textit{exploit}_1, \textit{apt8})$ -$\!\prec$ \\
				&            & \hskip6mm  \textsf{time\_diff}$(\textit{interval}, \textit{robotmafia})$ \\
			\end{tabular}
		}}
		\caption{A ground argumentation framework.}
		\label{fig:gndArgEx}
	\end{figure}

	\begin{definition}(\textbf{Argument}) An argument for a literal $L$ is a pair $\langle \ar, L \rangle$, where $\ar\subseteq\Pi$ provides a minimal proof for $L$ meeting the requirements: (1) $L$ is defeasibly derived from $\ar$\footnote{This means that there exists a derivation consisting of a sequence of rules that ends in $L$---that possibly includes defeasible rules.}, (2) $\Theta \cup \Omega \cup \Delta$ is not contradictory, and (3) $\ar$ is a minimal subset of $\Delta$ satisfying~1 and~2, denoted $\langle \ar, L \rangle$.
	\end{definition}	
	Literal $L$ is called the {\em conclusion} supported by the argument, and $\ar$ is the \emph{support}. 
	An argument $\langle \B, L\rangle$
	is a {\em subargument} of $\langle \ar, L'\rangle$ iff $\B \subseteq \ar$.
	The following examples show arguments for our scenario.

	\begin{figure}[t]
		\small
		\fbox{
			\parbox{0.98\columnwidth}{
				\begin{tabular}{ll}
					$\langle \ar_1$, \textsf{replay\_attack}$(\textit{exploit}_1$) $\rangle$& $\ar_1 = \{\delta_1, \theta_1,\theta_3\}$\\	
					$\langle \ar_2$, \textsf{deception}$(\textit{exploit}_1$, \textit{apt8}) $\rangle$& $\ar_2 = \{\delta_1,\delta_2,\theta_2\}$\\
					$\langle \ar_3$, \textsf{culprit}$(\textit{exploit}_1$, \textit{apt8})$\rangle$& $\ar_3 = \{\delta_1, \delta_2,\delta_3\}$\\
					$\langle \ar_4$, \textsf{$\neg$culprit}$(\textit{exploit}_1$, \textit{apt8})$\rangle$ & $\ar_4 = \{\delta_1,\delta_4,\theta_3\}$
				\end{tabular}
			}}
			\caption{Example ground arguments from Figure~\ref{fig:gndArgEx}.}
			\label{fig:gndArgsEx}
		\end{figure}

		\begin{example}
			\figurename~\ref{fig:gndArgsEx} shows example arguments based on the KB from \figurename~\ref{fig:gndArgEx}; 
			here, $\big \langle \ar_1$, \textsf{replay\_attack}$(\textit{exploit}_1)\big\rangle$ is a subargument of
			$\big \langle \ar_2$, \textsf{deception}$(\textit{exploit}_1, \textit{apt8})\big\rangle$  and
			$\big \langle \ar_3$, \textsf{culprit}$(\textit{exploit}_1, \textit{apt8})\big\rangle$.
		\end{example}
		
		For a given argument there may be counter-arguments that contradict it. For instance, referring to \figurename~\ref{fig:gndArgsEx}, we can see that $\ar_4$ attacks $\ar_3$. A {\em proper defeater} of an argument $\langle A, L\rangle$ is a counter-argument
		that---by some criterion---is considered to be better than $\langle \ar, L\rangle$; if the two are incomparable according
		to this criterion, the counterargument is said to be a {\em blocking} defeater. The default criterion used in DeLP for argument comparison is \textit{generalized specificity}~\cite{stolzenburg2003}.
		
		A sequence of arguments is called an \textit{argumentation line}. There can be more than one defeater argument, which leads to a tree structure that is built from the set of all argumentation lines rooted in the initial argument. In this \textit{dialectical tree}, every child can defeat its parent (except for the root), and the leaves represent unchallenged arguments; this creates a map of all possible argumentation lines that decide if an argument is defeated or not.
		Arguments that either have no attackers or all attackers have been defeated are said to be {\em warranted}.
		
		Given a literal $L$ and an argument $\big \langle \ar, L \big \rangle$, in order to decide whether or not a literal $L$ is warranted,
		every node in the dialectical tree $\Tree{ \langle \ar, L \rangle}$ is recursively marked as ``D'' (\textit{defeated}) or
		``U'' (\textit{undefeated}), obtaining a marked \textit{dialectical tree} $\MTree{ \langle \ar, L  \rangle}$ where:
		\begin{itemize}
			\item All leaves in $\MTree{ \langle \ar, L  \rangle}$ are marked as ``U''s, and
			
			\item Let $ \langle \B, q  \rangle$ be an inner node of $\MTree{ \langle \ar, L  \rangle}$.
			Then, $ \langle \B, q  \rangle$ will be marked as ``U''
			iff every child of $ \langle \B, q  \rangle$ is marked as ``D''.
			Node $ \langle \B, q  \rangle$ will be marked as ``D''
			iff it has at least one child marked as ``U''.
		\end{itemize}
		Given argument $\langle \ar, L \rangle$ over $\Pi$, if the root of $\MTree{ \langle \ar, L  \rangle}$ is marked ``U'', then
		$\MTree{ \langle \ar, h  \rangle}$ \textit{warrants} $L$ and that $L$ is \textit{warranted} from $\Pi$.
		(Warranted arguments correspond to those in the grounded extension of a Dung argumentation system \cite{dung1995}.)
		
		In practice, an implementation of DeLP accepts as input sets of facts, strict rules, and defeasible rules.  Note that while the set of facts and strict rules is consistent (non-contrdictory), the set of defeasible rules can be inconsistent.  
		We engineer our cyber-attribution framework as a set of defeasible and strict rules whose structure was created manually, but are dependent on values learned from a historical corpus of data.  Then, for a given incident, we instantiate a set of facts for that situation.  This information is then provided as input into a DeLP implementation that uses heuristics to generate all arguments for and against every possible culprit for the cyber attack.  Dialectical trees based on these arguments are analyzed, and a decision is made regarding which culprits are \textit{warranted}.  This results in a reduced set of potential culprits, which we then use as input into a classifier to obtain an attribution decision.

\section{Baseline Argumentation Model (BM)}
\label{lat}

In~\cite{fosint} machine learning techniques were leveraged on the CTF data to identify the attacker. We will now provide a summary of the results obtained. The experiment was performed as follows. The dataset was divided according to the target team, building 20 subsets, and all the attacks were then sorted according to time. 
The first 90\% of the attacks were reserved for training and the remaining 10\% for testing. The byte and instruction histograms were used as features to train and test the model. Models constructed using a random forest classifier performed the best, with an average accuracy of 0.37. Most of the misclassified samples tend to be deceptive attacks and their duplicates. 

When using machine learning approaches it is difficult to map the reasons why a particular attacker was predicted, especially in cases of deception where multiple attackers were associated with the same attack. Knowing the arguments that supported a particular decision would greatly aid the analyst in making better decisions dealing with uncertainty. To address this issue we now describe how we can form arguments/rules based on the latent variables computed from the training data, given an attack for attribution.

We use the following notation: let $\e$ be the test attack under consideration aimed at target team $\x$, $\y$ represent all the possible attacking teams, and $\dep$ be the set of all deceptive teams (those using the same payload to target the same team) if the given attack is deceptive in the training set. For non-deceptive attacks, $\dep$ will be empty. We note that facts cannot have variables, only constants (however, to compress the program for presentation purposes, we use {\em meta-variables} in facts). To begin, we define the facts:
$\theta_{1} = \textsf{attack ($\e, \x$)}, \theta_{2} = \textsf{first\_attack ($\e, \y$)},
\theta_3 =  \textsf{last\_attack ($\e, \y$)}$;
$\theta_1$ states that attack $\e$ was used to target team $\x$, $\theta_2$ states that team $\y$ was the first team to use the attack $\e$ in the training data, and similarly $\theta_3$ states that team $\y$ was the last team to use the attack $\e$ in the training data. The first and last attacking team may or may not be the same. 
We study the following three cases:

\noindent
\textbf{Case 1: Non-deceptive attacks.}
In non-deceptive attacks, only one team uses the payload to target other teams in the training data. It is easy to predict the attacker for these cases, since the search space only has one team.
To model this situation, we define a set of defeasible and strict rules.

\begin{figure}[t]
	\small
	\fbox{
		\parbox{0.96\columnwidth}{
			\begin{tabular}{ll}
				\small
				$\omega_{1}=$ & \textsf{culprit($\e, \y$) $\leftarrow$}
				\textsf{last\_attack($\e, \y$)}, \textsf{replay\_attack($\e$).}\\[1ex]
				$\delta_{1}=$ & \textsf{replay\_attack($\e$) -$\!\prec$}
				\textsf{attack($\e, \x$),} \textsf{last\_attack($\e, \y$).}
			\end{tabular}
		}}
		\caption{Defeasible and strict rule for non-deceptive attack.}
		\label{nonD}
	\end{figure}

	In \figurename~\ref{nonD}, defeasible rule $\delta_1$ checks whether the attack was replayed in the training data. Since it is a non-deceptive attack, it can only be replayed by the same team. The strict rule $\omega_1$ then puts forth an argument for the attacker (culprit) if the defeasible rule holds and there is no contradiction for it.

	\begin{figure}[t]
		\small
		\fbox{
			\parbox{0.96\columnwidth}{
				\begin{tabular}{ll}
					\multicolumn{2}{l}{$\theta_{1}=$ \textsf{decep ($\e, \x$)}, $\theta_{2}=$ \textsf{frequent ($\e, F$)}}\\
					\multicolumn{2}{c}{\rule{0.8\columnwidth}{0.4pt}} \\
					$\omega_{1}=$ & \textsf{$\neg$culprit($\e, \y$) $\leftarrow$} 
					\textsf{first\_attack($\e, \y$)}, \textsf{decep($\e, \x$)}\\
					$\omega_{2}=$ & \textsf{culprit($\e, F$) $\leftarrow$}
					\textsf{frequent($\e, F$)}, \textsf{deception ($\e, \dep_i$)}\\
					\multicolumn{2}{c}{\rule{0.8\columnwidth}{0.4pt}} \\
					$\delta_{1}=$ & \textsf{replay\_attack($\e$) -$\!\prec$}
					\textsf{attack($\e, \x$),} \textsf{last\_attack($\e, \y$)}\\
					$\delta_{2}=$ & \textsf{deception($\e, \dep_i$) -$\!\prec$} 
					\textsf{replay\_attack($\e$),} \\ 
					& \phantom{\textsf{deception($\e, \dep_i$) -$\!\prec$}} \textsf{first\_attack($\e, \y$)}\\
					$\delta_{3}=$ & \textsf{culprit($\e, \dep_i$) -$\!\prec$}
					\textsf{deception($\e, \dep_i$),} \textsf{first\_attack($\e, \y$)}
				\end{tabular}
			}}
			\caption{Facts and rules for deceptive attacks.}
			\label{deceptive}
			
		\end{figure}

		\noindent
		\textbf{Case 2: Deceptive attacks.}
		These attacks form the majority of the misclassified samples in~\cite{fosint}. The set $\dep$ is not empty for this case; let $\dep_i$ denote the deceptive teams in $\dep$. We also compute the most frequent attacker from the training data given a deceptive attack. Let the most frequent deceptive attacker be denoted as $F$. The DeLP components that model this case are shown in Figure~\ref{deceptive};
		fact $\theta_1$ indicates if the attack $\e$ was deceptive towards the team $\x$ and $\theta_2$ indicates the most frequent attacker team $F$ from the training set. The strict rule $\omega_1$ indicates that in case of deception the first team to attack (Y) is not the attacker, $\omega_2$ states that the attacker should be $F$ if the attack is deceptive and $F$ was the most frequent deceptive attacker. For the defeasible rules, $\delta_1$ deals with the case in which the attack $\e$ was replayed, $\delta_2$ deals with the case of deceptive teams from the set $\dep$, $\delta_3$ indicates that all the deceptive teams are likely to be the attackers in the absence of any contradictory information. and $\delta_4$ states that the attacker should be $F$ if the attack is deceptive and $F$ was the most frequent attacker.

		\noindent
		\textbf{Case 3: Previously Unseen Attacks.}
		The most difficult attacks to attribute in the dataset are the unseen ones, i.e. attacks first encountered in the test set and thus did not occur in the training set. To build constructs for this kind of attack we first compute the $k$ nearest neighbors from the training set according to a simple Euclidean distance between the byte and instruction histograms of the two attacks. In this case we choose $k = 3$. For each of the matching attacks from the training data we check if the attack is deceptive or non-deceptive. If non-deceptive, we follow the procedure for Case 1, otherwise we follow the procedure for Case 2. Since we replace one unseen attack with three seen attacks, the search space for the attacker increases for unseen attacks.\medskip
		
\begin{figure}[t]
			\centerline{\includegraphics[width=\columnwidth]{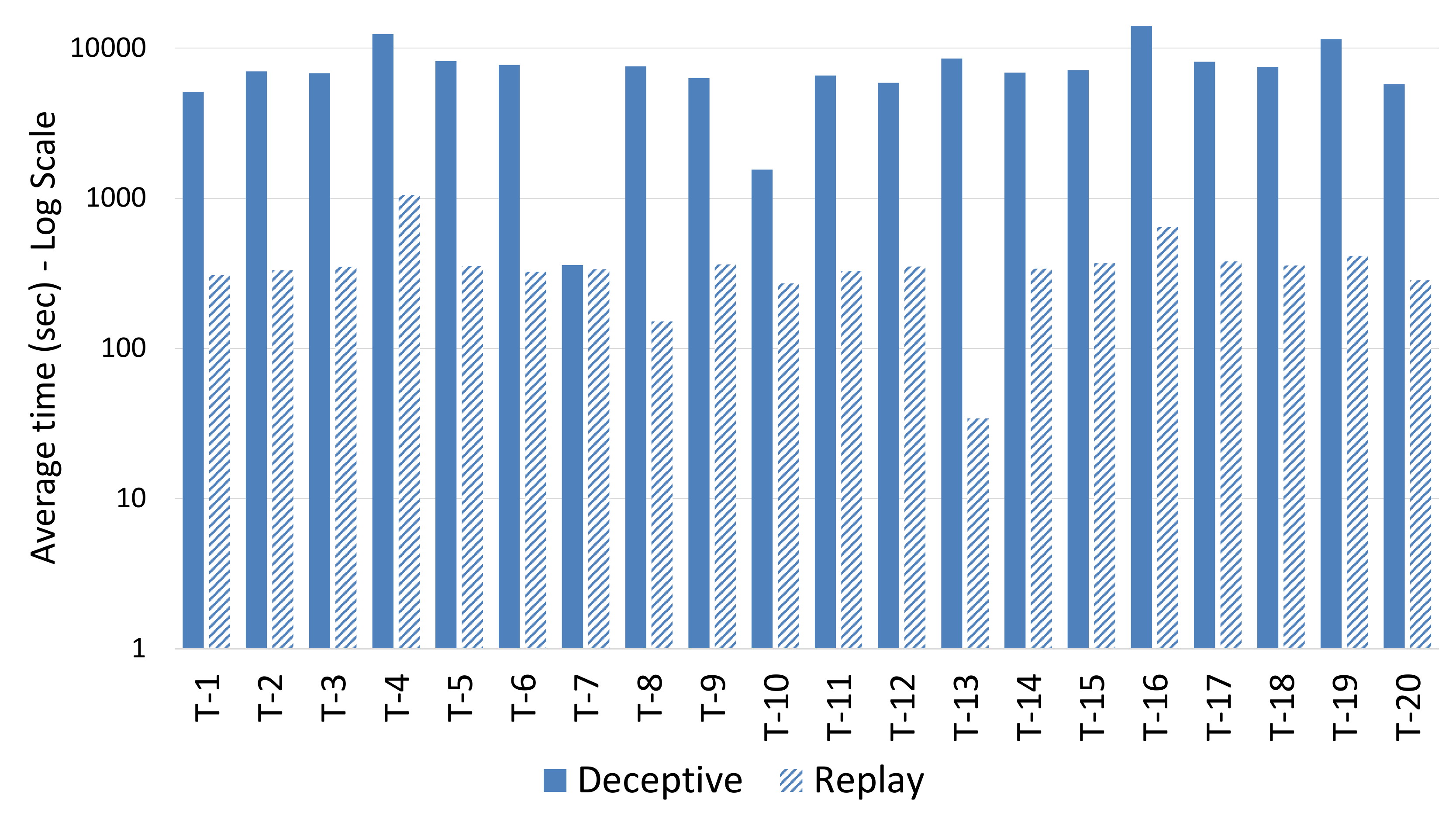}}
			\captionsetup{justification=centering}
			
			\caption{\textmd{Average time for team to perform a deceptive attack and replay its own attack (Log-scale).}}
		
			\label{timend}
		\end{figure}

		\noindent\textbf{Attacker Time Analysis:}
		The CTF data provides us with time stamps for the attacks in the competition. We can use this information to come up with rules for/against an argument for a team being the attacker. We compute the average time for a team to replay its own attack given that it was the first one to deploy the attack (see \figurename~\ref{timend}). It can be observed that teams like \textit{more smoked leet chicken} (T-13) and \textit{Wowhacker-bios} (T-8) are very quick to replay their own attacks as compared to other teams. \figurename~\ref{timend} also shows the average time for a team to perform a deceptive attack. Teams like \textit{The European} (T-7) and \textit{Blue lotus} (T-10) are quick to commit deception, while others take more time.
		
		We use this time information to narrow down our search space for possible attackers. In particular, for a deceptive test sample, we compute the time difference between the test sample and the training sample that last used the same payload. We denote this time difference as $\bigtriangleup t$, and include it as a fact $\theta_1$. We then divide the deceptive times from \figurename~\ref{timend} into appropriate intervals; each team is assigned to one of those time intervals. We then check which time interval $\bigtriangleup t$ belongs to and define a defeasible rule $\delta_1$ that makes a case for all teams not belonging to the interval to not be the culprits, as shown in \figurename~\ref{time}.
		
		\begin{figure}[t]
			\small
			\fbox{
				\parbox{0.96\columnwidth}{
					\begin{tabular}{lll}
						\small
						$\Theta:$& $\theta_{1}=$ & \textsf{timedifference ($\e, \x$)}\\[1ex]
						& &For $\y$ $\notin$ interval:\\
						$\Delta:$ &$\delta_{1}=$ & \textsf{$\neg$culprit($\e, \y$) -$\!\prec$}
						\textsf{timedifference ($\e, \x$).}
					\end{tabular}
				}}
				\caption{Time facts and rules. Interval indicates a small portion of the entire deceptive time (for instance $<$ 2000 sec, $>$ 8000 sec and so on).}
				\label{time}
			\end{figure}

			We now provide a summary of the experimental results---the setup is similar to~\cite{fosint}: the dataset is sorted by time for each target team, the first 90\% of the data is used for training and the remaining 10\% for testing. The constructs for all test samples based on the cases discussed in the previous section are computed, and these arguments are used as input to the DeLP implementation. For each test sample the DeLP system is queried to find all possible attackers (culprits) based on the arguments provided. If there is no way to decide between contradicting arguments, these are blocked and thus return no answers. Initially, the search space for each test sample is 19 teams (all except the one being attacked).
			
			After running the queries to return the set of possible culprits, the average search space across all target teams is 5.85 teams. This is a significant reduction in search space across all target teams; to gauge how much the reduced search space can aid an analyst in predicting the actual culprit, a metric is computed that checks if the reduced search space contains the ground truth (actual culprit). For all the target teams, the ground truth is present on average in almost 66\% of the samples with reduced search space.
			For some teams like \textit{more smoked leet chicken} (T-13) and \textit{raon\_ASRT (whois)} (T-17) the average reduced search space is as low as 1.82 and 2.9 teams, with high ground truth fraction of 0.69 and 0.63, respectively.
			
			Predictive analysis is then performed on the reduced search space. The experimental setup is similar to the one described earlier; the only difference this time is instead of having a 19 team search space as in~\cite{fosint}, the machine learning approach is allowed to make a prediction from the reduced search space only; a random forest is used for learning, which has been shown to have the best performance for CTF data~\cite{fosint}. 
			
			We report the following average accuracies across 20 target teams; the accuracy achieved after running Random forest without applying the argumentation-based techniques, as reported in~\cite{fosint}, is 0.37. This was the best performing approach using standard machine learning techniques. The baseline model achieves an average accuracy of 0.5, which is significantly better than the average accuracy of 0.37 in~\cite{fosint}.

\section{Extended Baseline Model \upperRomannumeral{1} (EB1)}
\label{eb1}
Previously Unseen attacks make up almost 20\% of the test samples for each target team. On analyzing the misclassification from the baseline argumentation model, we observe that the majority of the previously unseen attacks get misclassified ($>$80\%). The misclassifications can be attributed to two reasons:
(i) the reduced search space is not able to capture the ground truth for unseen attacks, leading the learning model to a wrong prediction; and (ii) we represent each unseen attack by the 3 most similar attacks in the training data; this leads to an increase in the search space and many choices for the learning model.

We address these issues by proposing two sets of defeasible rules. First, for each target team we compute from the training set the top 3 teams that come up with the most unique exploits, as these teams are more likely to launch an unseen attack in the test set. The intuition behind this rule is the fact that not all teams write their own exploits, most teams just capture a successful exploit launched by other teams and repackage it and use it as their own (deception). The second set of rules is proposed to avoid addition of less similar teams to the reduced search space. In the baseline model we use 3-nearest neighbors to represent an unseen attack. In this extended version we consider only the nearest neighbors that are less than a particular threshold value $T$, which is decided for each target team separately. So, each attack will be represented by $k \leq 3$ teams depending upon the threshold requirement. In addition to the baseline model rules, we propose the following rules for deceptive attacks.
Let $\mathcal{U}$ denote the set of teams with the three highest numbers of unique attacks in the training data. Also, let $\mathcal{N}$ denote the set of three most similar culprits for the given unseen attack. 

\begin{figure}[t]
	\small
	\fbox{
		\parbox{0.96\columnwidth}{
			\begin{tabular}{lll}
				\small
				& & For ($n_i$ $\in$ $\mathcal{N}$ and $sim < T$):\\
				$\Theta:$& $\theta_1=$ & \textsf{threshold($\e, T$)}\\
				& & For $u_i$ $in$ $\mathcal{U}$:\\
				& $\theta_{2}=$ & \textsf {unique($\e, u_i$)}\\
				\multicolumn{3}{c}{\rule{0.9\columnwidth}{0.4pt}} \\
				$\Delta:$	&$\delta_{1}=$ & \textsf{culprit($\e, u_i$) -$\!\prec$}
				\textsf{threshold($\e, T$)}\\
				& & For $u_i$ $\in$ $\mathcal{U}$:\\
				&$\delta_{2}=$ & \textsf{culprit($\e, u_i$) -$\!\prec$}
				\textsf{unique($\e, u_i$)} \\
				
			\end{tabular}
		}}
		\caption{Rules for unseen attacks.}
		\label{unseen}
	\end{figure}
		
	The extended model is shown in \figurename~\ref{unseen}; the fact $\theta_1$ indicates the teams present in $\mathcal{N}$ and whose similarity is less than a particular threshold $T$, and $\theta_2$ indicates if the team $u_i$ was one of most unique attackers from set $\mathcal{U}$. For the defeasible rules, $\delta_1$ makes use of the fact $\theta_1$ stating that the teams in $\mathcal{N}$ that satisfy the threshold condition are likely to be the culprits, and  $\delta_2$ indicates that if $u_i$ is a unique attacker then it can be the culprit unless contradictory information is available. $\mathcal{U}$ is independent of the test samples and will be the same for all unseen attacks given a target team.
	
	On the contrary, for each of the similar payloads (three or fewer) computed from the training data we check if the attack is deceptive or non-deceptive. If non-deceptive, we follow the procedure for Case 1, otherwise we follow the procedure for Case 2 stated in the baseline argumentation model.

	\noindent\textbf{Experiment:}
	We evaluate EB1 using an experimental setup similar to the one for the baseline argumentation model. We report the average reduced search space and prediction accuracy for both EB1 and baseline model to provide a comparison. EB1 performs better than the baseline with an average accuracy of 0.53 vs. 0.50, and significantly better than the machine learning model without argumentation that has an average accuracy of 0.37. The improvement in performance is due to the larger fraction of reduced search spaces with ground truth present in them. Also, the search space reduced from on average 6.07 teams to 5.025 (less teams to consider). The results are reported in Table~\ref{final} along with a comparison to the second extended baseline argumentation model (EB2).

\section{Extended Baseline Model \upperRomannumeral{2} (EB2)}
\label{eb2}
Another source of misclassification in the baseline argumentation model is the presence of unseen deceptive teams and their duplicates. These refer to teams that did not use the exploit in the training set but started using it in the test set. It is difficult for a machine learning approach to predict such a team as being the culprit if it has not encountered it using the exploit in the training set. In our dataset these attacks comprise 15\% of the total, and up to 20\% for some teams.

\begin{figure}[t]
	\small
	\fbox{
		\parbox{0.96\columnwidth}{
			\begin{tabular}{lll}
				\small
				$\Theta:$& $\theta_{1}=$ & \textsf{timedifference ($\e, \x$)}\\
				
				\multicolumn{3}{c}{\rule{0.9\columnwidth}{0.4pt}} \\
				& &For $\y$ $\in$ interval:\\
				$\Delta:$ &$\delta_{1}=$ & \textsf{culprit($\e, \y$) -$\!\prec$}
				\textsf{timedifference ($\e, \x$).} \\
			\end{tabular}
		}}
		\caption{Time facts and rules. Interval indicates a small portion of the entire deceptive time (for instance $<$ 2000 sec, $>$ 8000 sec and so on).}
		\label{newtime}
	\end{figure}
	
	In order to address this issue we propose an extension of EB1, where we group together teams that have similar deceptive behavior based on the time information available to us from the training set; for instance teams that are deceptive within a certain interval of time (e.g., less than 2,000 secs.) after the first attack has been played are grouped together. For a given test attack we compute the time difference between the test attack and the last time the attack was used in the training set. We then assign this time difference to a specific group based on which interval the time difference falls in. In order to fine tune the time intervals, instead of using the average deceptive times averaged across all target teams (as used in the baseline model), we compute and use deceptive times for each target team separately. We model the time rules as stated in \figurename~\ref{newtime};
	fact $\theta_1$ states the time difference between the test sample and the last training sample to use that attack, defeasible rule $\delta_1$ on the other hand states that teams belonging to that interval (in which the time difference lies) are likely to be the culprits unless a contradiction is present. It is clear that this rule will increase the search space for the test sample, as additional teams are now being added as likely culprits.
	We observe that for EB2 the search space is increased by an average of almost 2.5 teams per test sample from EB1; at the same time the presence of ground truth in the reduced search space increased to 0.78, which is a significant improvement over 0.68.
	
	\noindent\textbf{Experiment:}
	We evaluate EB2 using an experimental setup similar to the one discussed in the baseline argumentation model. We report the prediction accuracies for each of the proposed baseline argumentation models for each of the target teams and compare it with the previous accuracy reported in~\cite{fosint}, denoted as ML. In Table~\ref{final} the second extended baseline model (EB2) performs the best with an average prediction accuracy of 62\% as compared to other proposed methods. The additions of teams based on time rules not only benefits detection of unseen deceptive teams but it also helps in predicting attackers for unseen attacks. The major reason for the jump in performance is that for most unseen deceptive team samples, the time rules proposed in the baseline model block all deceptive teams from being the culprit, leading to an empty set of culprits. The new set of rules proposed in EB2 adds similar-behaving teams to this set based on time information; the learning algorithm can then predict the right one from this set.
	\vspace{-1em}
	\begin{table}[h!]
		\caption{\textmd{Results Summary}}
		\footnotesize
		\label{final}
		\centering
		\renewcommand{\arraystretch}{1.5}
		
		\begin{tabular}{|p{1cm}|p{1cm}|p{1cm}|p{1cm}|p{1cm}| }
			\hline
			{\bf Team} &  {\bf ML~\cite{fosint}} & {\bf BM}& {\bf EB1}& {\bf EB2}\\ \hline
			
			\textsf{T-1} & 0.45 & 0.51 & 0.52 & \textbf{0.60} \\ \hline
			\textsf{T-2} & 0.22 & \textbf{0.45} & 0.38  & 0.43\\ \hline
			\textsf{T-3} & 0.30& 0.40 & 0.47  & \textbf{0.66}\\ \hline
			\textsf{T-4} & 0.26 & 0.44 & 0.42  & \textbf{0.44}\\ \hline
			\textsf{T-5} & 0.26 & 0.45 & 0.45 & \textbf{0.56}\\ \hline
			\textsf{T-6} & 0.5 & 0.49 & 0.55 & \textbf{0.7}\\ \hline
			\textsf{T-7} & 0.45 & 0.53 & 0.56 & \textbf{0.66}\\ \hline
			\textsf{T-8} & 0.42 & 0.61 & 0.58 & \textbf{0.74}\\ \hline
			\textsf{T-9} & 0.41 & 0.50 & 0.53 & \textbf{0.76}\\ \hline
			\textsf{T-10} & 0.30 & 0.42 & \textbf{0.41} & \textbf{0.41}\\ \hline
			\textsf{T-11} & 0.37 & 0.44 & 0.5 & \textbf{0.73}\\ \hline
			\textsf{T-12} & 0.24 & 0.43 & 0.36  & \textbf{0.52}\\ \hline
			\textsf{T-13} & 0.35 & 0.63 & 0.64 & \textbf{0.75}\\ \hline
			\textsf{T-14} & 0.42 & 0.52 & 0.53 & \textbf{0.67}\\ \hline
			\textsf{T-15} & 0.30 & 0.38 & 0.55 & \textbf{0.64}\\ \hline
			\textsf{T-16} & 0.43 & 0.48 & 0.55 & \textbf{0.65}\\ \hline
			\textsf{T-17} & 0.42 & 0.58 & 0.58 & \textbf{0.68}\\ \hline
			\textsf{T-18} & 0.48 & 0.50 & 0.52 & \textbf{0.65}\\ \hline
			\textsf{T-19} & 0.41 & 0.51 & 0.56 & \textbf{0.68}\\ \hline
			\textsf{T-20} & 0.48 & 0.51 & 0.64 & \textbf{0.71}\\ \hline
			
		\end{tabular}
		
	\end{table}

\vspace{-1em}
\section{Conclusion}
\label{con}

In this paper we demonstrated how an argumentation-based framework (DeLP) can be leveraged to improve cyber-attribution decisions by building DeLP programs based on CTF data; this affords a reduction of the set of potential culprits and thus greater accuracy when using a classifier for cyber attribution.  We are currently looking at implementing a probabilistic variant of DeLP~\cite{shakarian14}, as well as designing our own CTF event in order to better mimic real-world scenarios. Our new CTF will encourage deceptive behavior among the participants, and we are also enhancing our instrumentation of the CTF, allowing for additional data collection (host data is of particular interest).

\vspace{-0.2em}
\section{ Acknowledgments}
Authors of this work were supported by the U.S. Department of the Navy, Office of Naval Research, grant N00014-15-1-2742 as well as the Arizona State University Global Security Initiative (GSI) and by CONICET and Universidad Nacional del Sur, Argentina.
\bibliographystyle{abbrv}

\bibliography{ref}

\end{document}